\documentclass[journal]{IEEEtran}

\ifCLASSINFOpdf
\else
\fi
\usepackage[cmex10]{amsmath}
\usepackage{mdwmath}
\usepackage{mdwtab}
\usepackage{mathrsfs}
\usepackage[tight,footnotesize]{subfigure}
\usepackage{graphicx}
\usepackage{amsfonts}
\usepackage[font=small,
   justification=justified,
   format=plain]{caption} 
\usepackage{algorithm}% http://ctan.org/pkg/algorithm
\usepackage{algpseudocode}% http://ctan.org/pkg/algorithmicx
\usepackage{amssymb}
\usepackage{xcolor}
\usepackage{array}
\usepackage{bm}
\usepackage{subfigure}
\usepackage{enumitem}
\usepackage{url}

\newcommand{\z}[2]{z_{#1, #2}}

\newcommand{\zVec}[3]{\mathbf{z}_{#1, #2:#3}}
\newcommand{\xVec}[3]{\mathbf{x}_{#1, #2:#3}}
\newcommand{\mb}{\mathbf}

\begin{document}
 
\title{A Deep Learning Approach Towards Generating High-fidelity Diverse Synthetic Battery Datasets}

\author{Janamejaya Channegowda,~\IEEEmembership{Member,~IEEE,} Vageesh Maiya,~\IEEEmembership{Student Member,~IEEE,} and  Chaitanya Lingaraj ~\IEEEmembership{Member,~IEEE,}

         % <-this % stops a space
 }

\markboth{}%
{Shell \MakeLowercase{\textit{et al.}}:}
 
\maketitle

\begin{abstract}
Recent surge in the number of Electric Vehicles have created a need to develop inexpensive energy-dense Battery Storage Systems. Many countries across the planet have put in place concrete measures to  reduce and subsequently limit the number of vehicles powered by fossil fuels. Lithium-ion based batteries are presently dominating the electric automotive sector. Energy research efforts are also focussed on accurate computation of State-of-Charge of such batteries to provide reliable vehicle range estimates. Although such estimation algorithms provide precise estimates, all such techniques available in literature presume availability of superior quality battery datasets. In reality, gaining access to proprietary battery usage datasets is very tough for battery scientists. Moreover, open access datasets lack the diverse battery charge/discharge patterns needed to build generalized models. Curating battery measurement data is time consuming and needs expensive equipment. To surmount such limited data scenarios, we introduce few Deep Learning-based methods to synthesize high-fidelity battery datasets, these augmented synthetic datasets will help battery researchers build better estimation models in the presence of limited data. We have released the code and dataset used in the present approach to generate synthetic data. The battery data augmentation techniques introduced here will alleviate limited battery dataset challenges.
\end{abstract}

\begin{IEEEkeywords}
Lithium Ion batteries, Energy Storage, Synthetic Data, Deep Learning
\end{IEEEkeywords}

\IEEEpeerreviewmaketitle

\section{Introduction} 

Driven by the pressing need to curb life threatening pollutants emitted by the transportation domain, there has been a massive push by industries and state heads to electrify various transportation modes. Electric Vehicles (EVs) have garnered significant interest as a reliable replacement of fossil fuel vehicles. Li-ion batteries (LiBs) have a proven property of being energy dense with minimal self discharge \cite{batt_1,batt_2}. The Battery Management System (BMS), included  in the battery pack, is provided to protect the batteries \cite{batt_3}. State-of-Charge (SOC) of batteries is a crucial metric to gauge the remaining charge present in the battery pack. SOC is expressed as the  percentage of charge remaining in the battery to the maximum available battery capacity \cite{batt_4}. To provide dependable mileage of the EV, computing SOC is crucial. A more  straightforward measurement of SOC is difficult and in most cases SOC is determined from other battery parameters (Voltage, Temperature and Current) \cite{batt_5}. Precise SOC estimates rely on curated battery dataset.

\subsection{Literature Review}

The SOC determination techniques  available in literature can be classified into the following categories \cite{batt_6,batt_7,batt_8,batt_9}:
\begin{itemize}
\item Coulomb Counting Approach
\item Open Circuit Voltage (OCV) method
\item Physics model-based computation
\item Data-driven models
\end{itemize}
This paper is focused on the data-driven model approach, a new paradigm, which works with massive amounts of battery dataset \cite{batt_9}. The key difference between data-driven and other battery models is the reliance on high-quality curated dataset. The models rely on the inference provided by machine learning or deep learning techniques. The recent growth of such learning techniques has enabled battery researchers to learn and highlight the interdependencies between variables of interest. There have been few State-of-Charge estimators discussed in literature \cite{batt_10,batt_11}. Data-driven techniques greatly differ from conventional SOC estimators with the computation being performed with a uniform set of neural network parameters.

Although such data-driven techniques are known to perform exceptionally well as SOC estimators, they are not devoid of implementation complexity. Some of the issues include, scalability problems and the assumption that all train, test and validation dataset originate from identical distributions. This limits the generalizability of these algorithms. A Deep Neural Network (DNN) is an augmented variant of the Artificial Neural Network (ANN), which means a DNN  has much more layers compared to  ANN.  DNN has performed exceptionally well in the area of Computer Vision \cite{batt_23} and language translation \cite{batt_25}. However, deep neural networks have not been extensively explored for SOC estimation. Use of Long Short-Term Memory (LSTM) network have been trained and performance has been evaluated on various vehicle drive cycle data \cite{batt_26}. A Multilayer DNN has also been used for SOC prediction \cite{batt_27}. DNN has been a promising methodology which has been employed towards SOC estimation due to its ability to work with non-linear input data \cite{batt_28,falak}. All data-driven models assume access to completely labelled datasets but, privacy concerns limit the access of proprietary datasets to battery researchers. Limited dataset greatly restrict the advancement of precise battery degradation and SOC estimation algorithms. 
%The chief objectives of this paper are as follows:

\subsection{Motivation Behind Current Research Trajectory}
Most deep learning techniques heavily depend on superior quality labelled datasets to provide useful estimates. In most cases such dataset is unavailable due to the expense and time involved in gathering such dataset. In literature, most research efforts have  focussed on deriving insights from \textbf{existing} rudimentary open access datasets, in reality, manufacturers rarely share proprietary datasets with researchers. The deep learning methods mentioned here help to generate reliable synthetic battery parameter datasets. The produced data possess a high degree of similarity with the generated data. This paper addresses the key challenge of sparse datasets. The synthetic data greatly simplifies computational effort and saves experimentation expenses.

\subsection{Contributions}
Some of the key contributions in tackling the issue of sparse datasets has been listed here:
%\vspace*{-0.1cm}
\begin{enumerate}
\item A deep learning based synthetic data generation technique is introduced to surmount sparse data challenges
\item This work provides comprehensive comparison of various state-of-the-art deep learning techniques, relevant to time series data, to produce high-fidelity heterogeneous battery datasets
\item The goal is to release code and dataset used in this present approach to enable researchers reproduce and build upon our results
\end{enumerate}
The paper is organised as follows, fundamentals of deep learning architecture employed to produce synthetic data is described in Section II, details of the experiments performed to evaluate the synthetic data are provided in Section III. Results are discussed and inferences are elucidated in Section III followed by conclusion in Section IV.

\begin{figure}
\begin{center}
	\centerline{\includegraphics[width=\columnwidth]{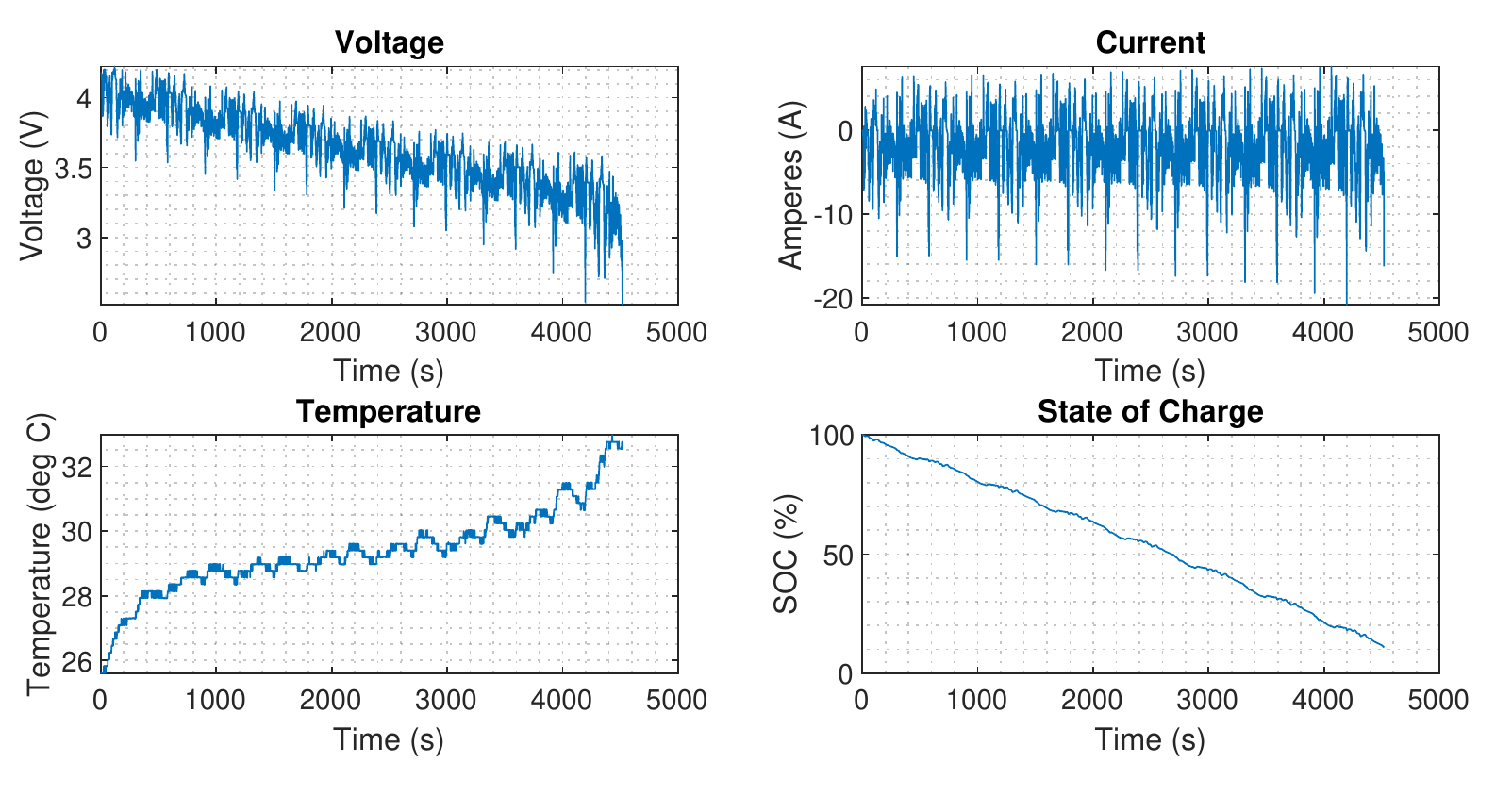}}
	\caption{Typical battery discharge characteristics of a Panasonic 18650 cell \cite{pana}}
	\label{fig:battt}
\end{center}
\end{figure} 

\begin{figure}
\begin{center}
	\centerline{\includegraphics[width=\columnwidth]{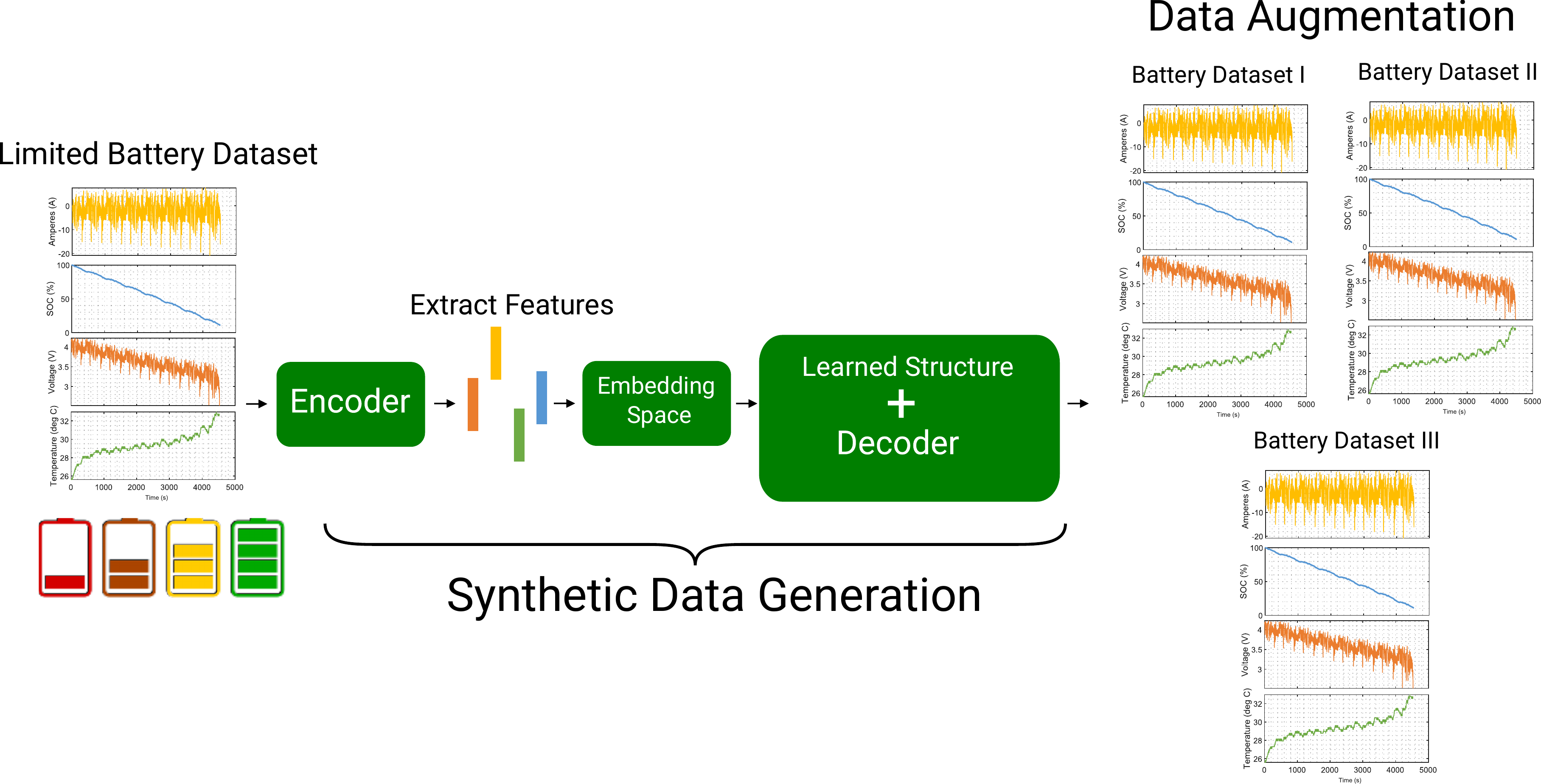}}
	\caption{An illustration of basic synthetic data generation methodology followed in any data-driven technique}
	\label{fig:block}
\end{center}
\end{figure} 

\section{Deep Learning-based Synthetic Data Generators}
Battery measurement datasets typically contain variations of key battery parameters such as Voltage, Current, Temperature and SOC captured over multiple charge and discharge cycles. One such discharge profile for an 18650 battery is shown in Fig.~\ref{fig:battt}. Battery measurement data collected over several cycles will enable researchers to devise accurate methodologies to estimate battery capacity fade over the life cycle of the cell.

Lack of such curated dataset warrants the use of synthetic data generators. Any synthetic data generation technique follows an encoder-decoder architecture style as illustrated in Fig.~\ref{fig:block}. The end goal in all such methods is to perform data augmentation, which helps to increase model training data by producing similar copies of groundtruth values. There has been significant progress in developing accurate deep learning time series based estimation approaches over the past few years, we evaluate the generation of synthetic data on primarily three state-of-the-art methods as listed in this section.

\subsection{Autoregressive Recurrent Networks Estimation (DeepAR)}

\begin{figure}
\begin{center}
	\centerline{\includegraphics[width=\columnwidth]{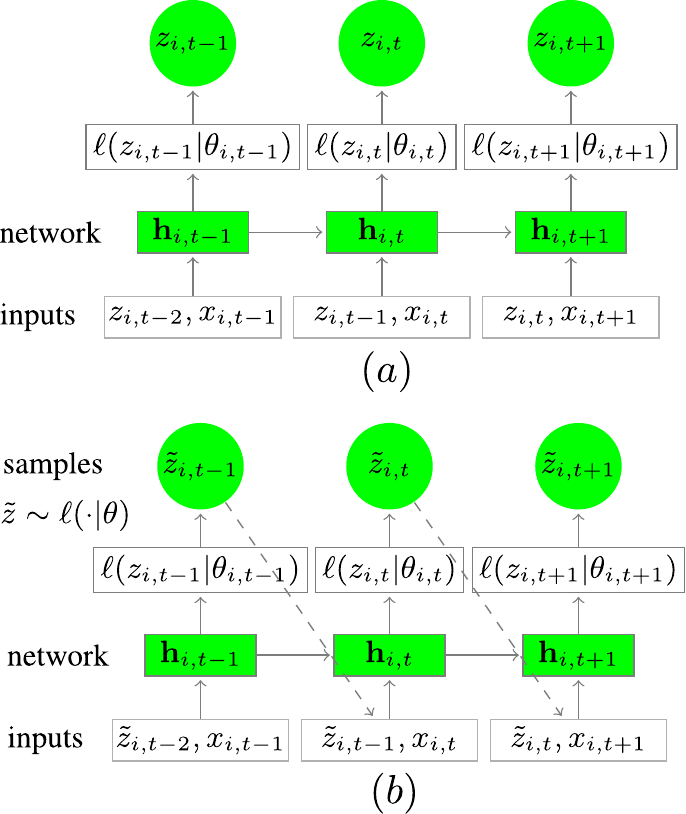}}
	\caption{Training procedure is depicted in (a), the inputs to the model are the temporal info $t$, data covariates $x_{i,t}$, historical targets $z_{i,t-1}$ and past outputs $\mathbf{h}_{i, t-1}$. Subsequently, the network result $\mathbf{h}_{i,t} = h(\mathbf{h}_{i,t-1}, z_{i,t-1}, \mathbf{x}_{i,t}, \Theta)$ is employed to compute $\theta_{i,t} = \theta(\mathbf{h}_{i,t}, \Theta)$ of  $\ell(z|\theta)$. The estimation methodology is shown in (b), the input in this case is past data $z_{i,t}$ in the duration $t<t_0$. For $t\ge t_0$ a trial  $\hat{z}_{i,t} \sim \ell(\cdot|\theta_{i,t})$  which is provided to the next iteration. This process is continued until the end of the estimated time series $t=t_0 + T$, thus generating a synthetic sequence into the future \cite{deepar}}
	\label{fig:deepar}
\end{center}
\end{figure} 

Multiple research attempts have been made in the past to employ fundamental neural network based architectures for forecasting future datapoints \cite{ForecastingNNSOA},\cite{Gers2001}. Recently Nikolaos et al \cite{Kourentzes2013} used focussed forecasting by using intermittent datapoints but with unsatisfactory results. Most of the initial work involved univariate  data and model fit was evaluated on such time series data separately \cite{kaastra1996designing,ghiassi2005dynamic}. These models till date have not been employed for data augmentation purposes.

Among such models Recurrent Neural Networks have been found to be very useful, especially in signal processing and natural language processing applications \cite{graves2013,sutskever2014}. The most useful features of long-sequence forecasting \cite{deepar} that are relevant to synthetic time-series data are as follows:
\begin{itemize}
\item Overall approximate prediction distribution is considered to obtain accurate variable estimates
\item This approach uses a negative binomial likelihood to provide better estimates of variables of interest
\end{itemize}

The entire procedure is shown in Fig.~\ref{fig:deepar} which shows the training \& estimation methods. The final goal is to derive the conditional $z_{i,t}$ ($i$ is the sequence under consideration at instant $t$) as shown in Equation \ref{eq:condDist}.

\begin{equation}
    P(\mathbf{z}_{i,t_0:T} | \mathbf{z}_{i,1:t_{0-1}},\mathbf{x}_{i,1:T})
    \label{eq:condDist}
\end{equation}

The distribution shown in Equation \ref{eq:condDist} is tied with the series $[\z{i}{t_0}, \z{i}{t_0 + 1}, \ldots, \z{i}{T}] := \zVec{i}{t_0}{T}$, in this case, it is assumed that the earlier values are known (\hbox{$[\z{i}{1}, \ldots, \z{i}{t_0-2}, \z{i}{t_0-1}] := \zVec{i}{1}{t_0-1}$}), we also consider the covariates ($\xVec{i}{1}{T}$) to be known for every time step. In this approach $[t_0, T]$ is the prediction window and $[1, t_0-1]$ is the conditioning window. The training period guarantees that both window ranges are in the past and $\z{i}{t}$ is perceived. The predicted sequence, $\z{i}{t}$, exists only in the conditioning interval. The entire model is based on the recurrent neural network architecture \cite{graves2013,sutskever2014}. The foundation of the network means that the value at the very last step $\z{i}{t-1}$ is provided as input and also as the recurrent value. The autoregressive portion implies that the previous neural network output ($\mathbf{h}_{i,t-1}$) is provided as input towards the next time step. In the experiments conducted on the battery datasets in this paper, the autoregressive recurrent network was employed for conditioning and prediction interval.

\subsection{Neural Basis Expansion Analysis (N-BEATS)}
\begin{figure}[t!]
\centering
	\centerline{\includegraphics[width=\columnwidth]{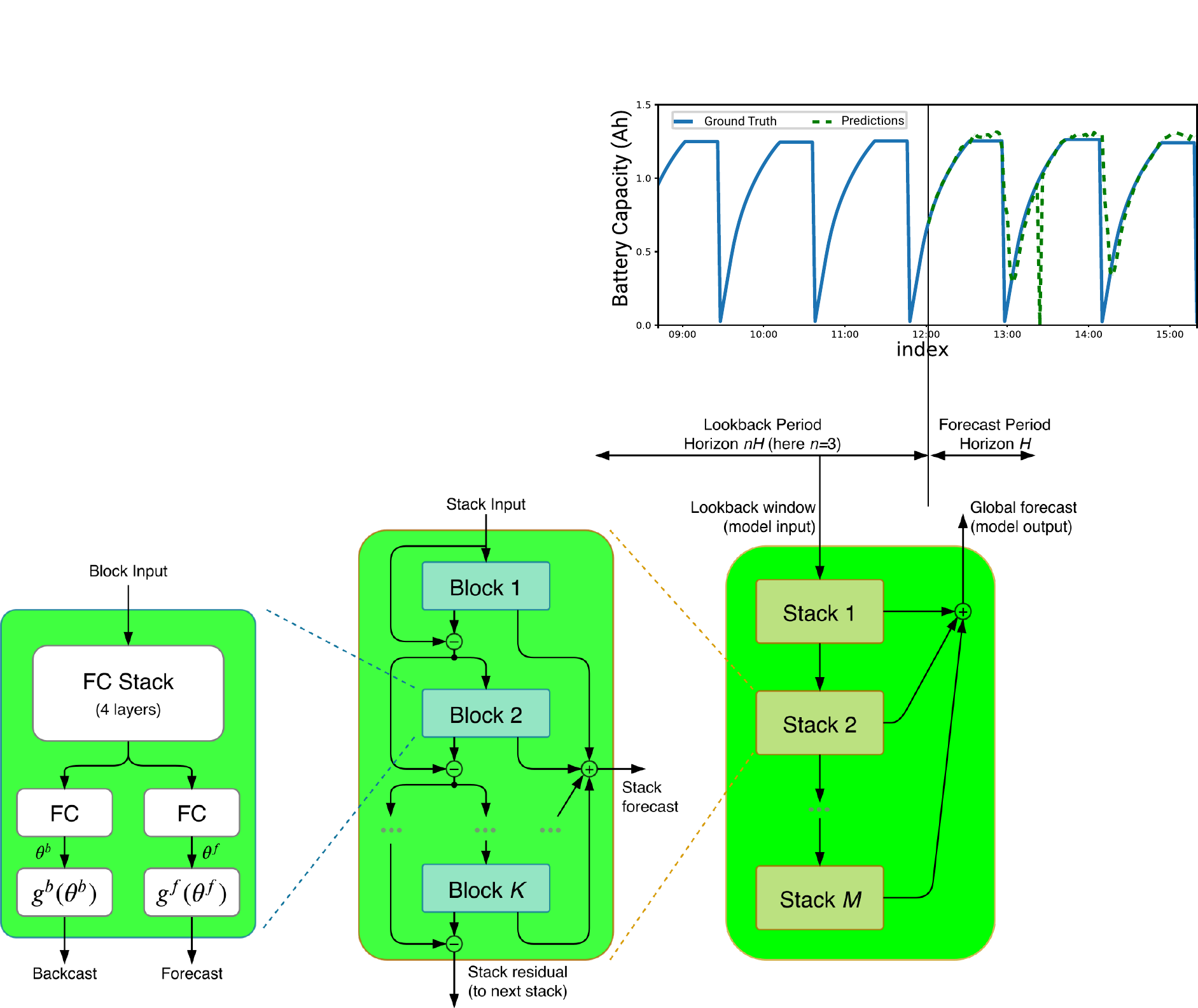}}
\caption{Basic blocks of the network used in Neural basis expansion analysis. The calculation of forward ($\theta^f$) \& backward ($\theta^b$) coefficients is performed. A single stack has many layers which have similar $g^b$ \& $g^f$ between them.}
\label{fig:nbeats}
\end{figure}

Another recent deep neural architecture is the Neural basis expansion analysis approach for long-sequence time-series forecasting \cite{nbeats}. The entire architecture is shown in Fig.~\ref{fig:nbeats}. The model takes as input $\vec{x}$ and the output being $\widehat{\vec{x}}$ \& $\widehat{\vec{y}}$. The input, $\vec{x}$, consists of a  historical window, the total sequence length of this window can contain many forecast horizons $H$. The estimated values, $\widehat{\vec{y}}$, contains the estimated length $H$ length \& $\widehat{\vec{x}}$. The primary blocks of this architecture are:

\begin{itemize}
\item A fully connected network generating the forward predictor ($\theta^f$) \& backward predictor ($\theta^b$)
\item The Forward layer  ($g^f$) \& the backward layer ($g^b$)
\end{itemize}
The Fully Connected ($FC$) layers are described by the following equations:
\begin{align}
\vec{h}_1 = FC_{1}(\vec{x}) \\
\vec{h}_{2} = FC_{2}(\vec{h}_{1}) \\
\vec{h}_{3} = FC_{3}(\vec{h}_{2}) \\
\vec{h}_{4} = FC_{4}(\vec{h}_{3}) \\
\theta^b = Linear^{b}(\vec{h}_{4}) \\
\theta^f = Linear^{f}(\vec{h}_{4})
\end{align} 
Here, $Linear$ depicts the projection layer, this signifies that $\theta^f = \vec{W}^{f} \vec{h}_{4}$, the entire purpose of this module is to calculate coefficients $\theta^f$, this coefficient enables optimization of the estimates $\widehat{\vec{y}}$. Ultimately, the coefficients $\theta^b$ are provided as inputs to the backward basis layer ($g^b$) to produce the estimate of $\vec{x}$. The next block of the architecture links $\theta^f$ \& $\theta^b$  to the estimates throughout the layers $\widehat{\vec{y}} = g^f(\theta^f)$ \& $\widehat{\vec{x}} = g^b(\theta^b_{\ell})$, the equation describing these operations are listed in Equation~\ref{equation:nbeats}.
\begin{equation}\label{equation:nbeats} 
\widehat{\vec{y}}_{\ell} = \sum_{i=1}^{\dim(\theta^f_{\ell})} \theta^f_{\ell,i} \vec{v}^f_{i}, \quad  \widehat{\vec{x}}_{\ell} = \sum_{i=1}^{\dim(\theta^b_{\ell})} \theta^b_{\ell,i} \vec{v}^b_{i}.
\end{equation}

\begin{figure}
\begin{center}
	\centerline{\includegraphics[width=\columnwidth]{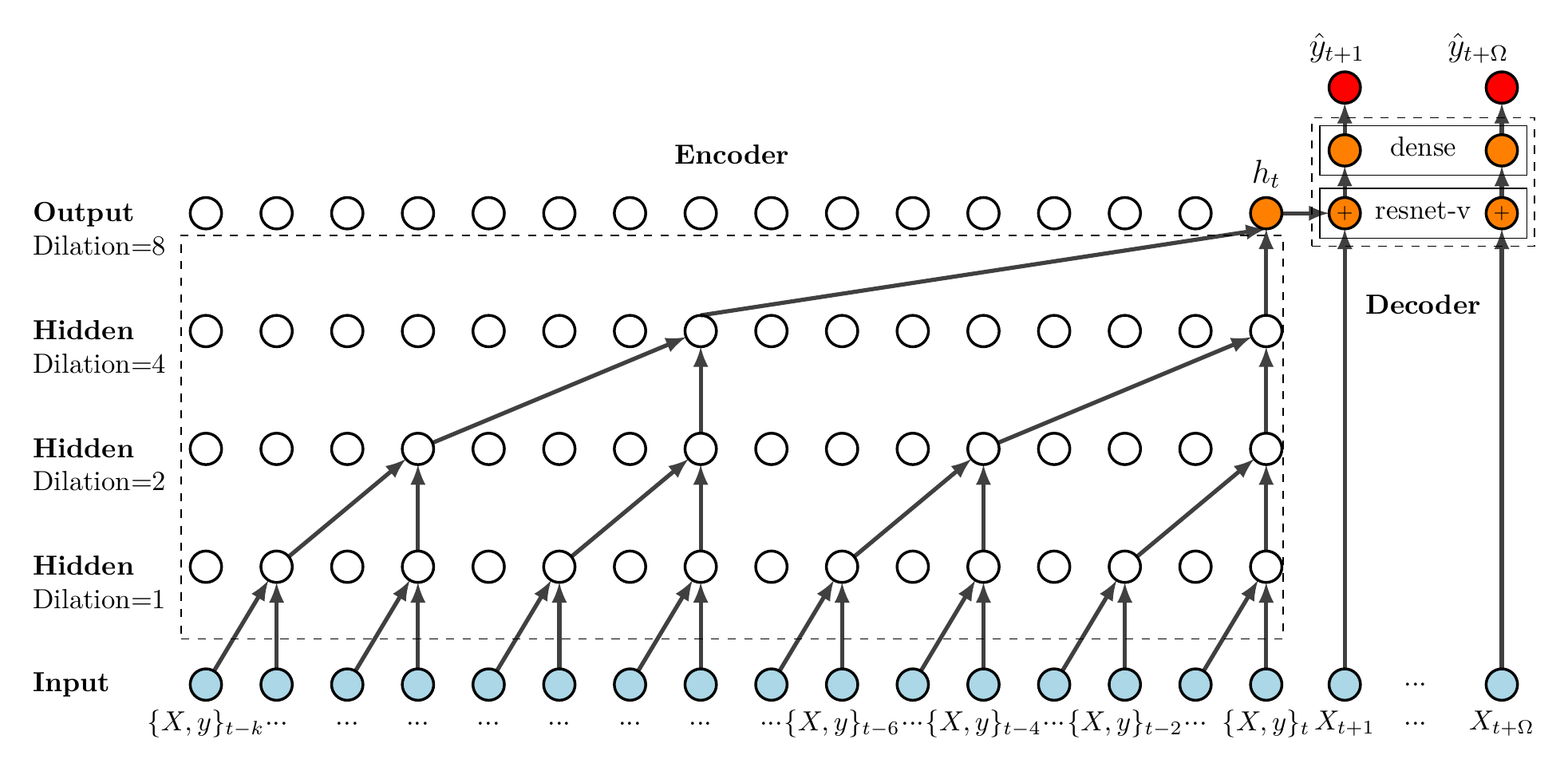}}
	\caption{DeepTCN algorithm, the encoder consists of stacked dilated nets needed to embed all the long sequence temporal information. The decoder has a variant of residual block \& an output layer which is dense. The dense layer links the output of the residual block to the final battery parameter estimates.}
	\label{fig:tcn_dcc}
\end{center}
\end{figure} 

\subsection{Deep Temporal Convolutional Network (DeepTCN)}

\begin{figure}
\begin{center}
	\centerline{\includegraphics[width=\columnwidth]{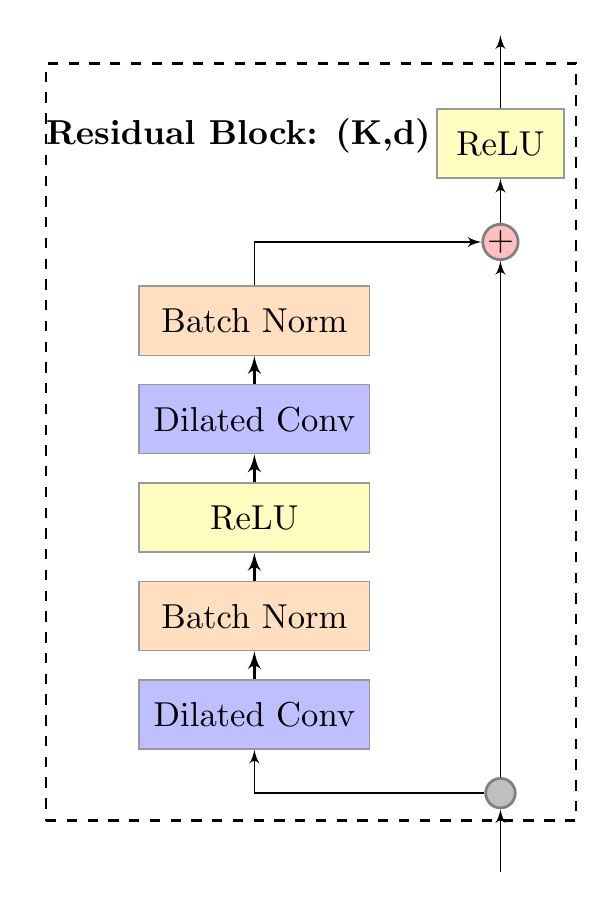}}
	\caption{Encoder of DeepTCN is shown here, every residual module has many layers of convolutions, the very first layer is led by batch normalization \& ReLU. In the encoder the output is fed as the input to the residual block and then by a ReLU activation function.}
	\label{fig:encoder_tcn_resnet}
\end{center}
\end{figure}

\begin{figure}
\begin{center}
	\centerline{\includegraphics[width=\columnwidth]{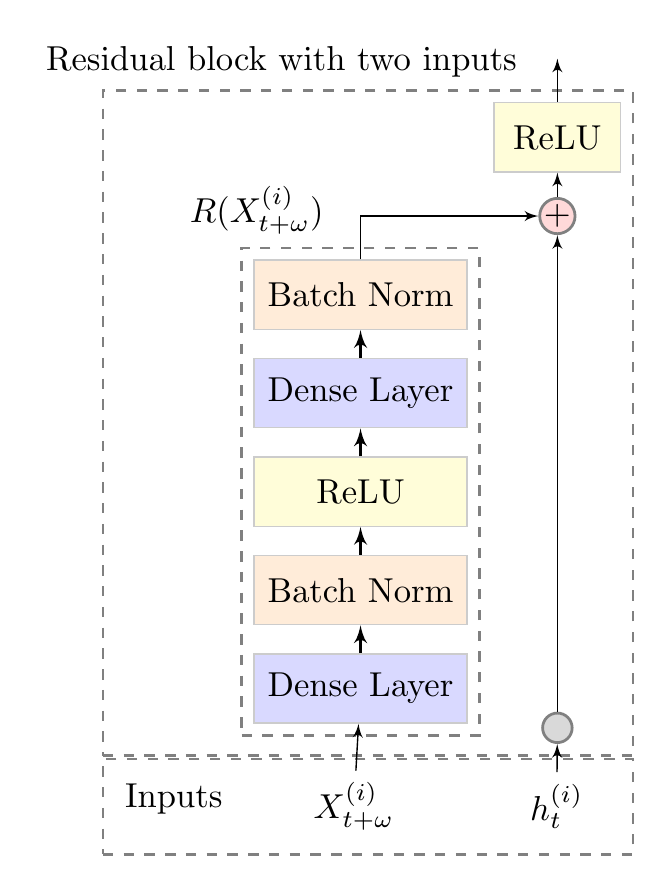}}
	\caption{Decoder of DeepTCN, the final decoder module output is $h_{t}^{(i)}$, here the estimated covariates are represented as $X_{t+\omega}^{(i)}$. A nonlinear function ($R(\cdot)$) is imposed on $X_{t+\omega}^{(i)}$. Finally in the residual function $R(\cdot)$, a batch normalization along with a dense layer is applied to predict the estimated covariates. Finally a ReLU activation function is applied by one more dense layer \& batch normalization.}
	\label{fig:decoder_tcn_futureResnet}
\end{center}
\end{figure}

The Deep Temporal Convolutional Network (DeepTCN), is a non-autoregressive estimation technique, this approach has proved to be particularly useful due to the following advantages \cite{deeptcn}:

\begin{enumerate}
\item Furnishes parametric \& non-parametric means to estimate probability density
\item Ability to learn dormant correlation between multivariate series which equips it to manage advanced battery datasets
\item Provides flexible means to include additional variables during estimation phase
\end{enumerate}
The complete architecture of DeepTCN is shown in Fig.~\ref{fig:tcn_dcc}, the encoder and decoder portions are illustrated in Fig.~\ref{fig:encoder_tcn_resnet} and Fig.~\ref{fig:decoder_tcn_futureResnet} respectively. In any time series estimation procedure the series can be described as ${\mathbf y}_{1:t} = \{y_{1:t}^{(i)}\}_{i=1}^N$, the estimated series is denoted by ${\mathbf y}_{(t+1):(t+\Omega)} = \{y_{(t+1):(t+\Omega)}^{(i)}\}_{i=1}^N$, here $N$ denotes the total number of time series, $t$ denotes past datapoints \& $\Omega$ is the estimation sequence. The chief aim is to estimate the entire distribution of future datapoints $P\left({\mathbf y}_{(t+1):(t+\Omega)}|{{\mathbf y}_{1:t}}\right)$.

The conventional generative architectures employed in time series analysis consider the joint probability of subsequent datapoints provided that we have access to historical data as depicted in the Equation.~\ref{cd1}.
\begin{equation}\label{cd1}
    P\left({\mathbf y}_{(t+1):(t+\Omega)}|{\mathbf y}_{1:t}\right) = \prod\limits_{\omega=1}^{\Omega} p({\mathbf y}_{t+\omega}|{\mathbf y}_{1:t+\omega-1}),
\end{equation}
Typically, generative models do not generalize well in all real-world applications, there is also accumulation of error for each prediction sequence, this is due to the fact that every estimated value is fed back as raw data to provide longer horizon estimates. In DeepTCN, the joint distribution of all estimates are observed right away. 
\begin{equation}\label{direct}
    P\left({\mathbf y}_{(t+1):(t+\Omega)}|{\mathbf y}_{1:t}\right) =
    \prod\limits_{\omega=1}^{\Omega} p({\mathbf y}_{t+\omega}|{\mathbf y}_{1:t}).
\end{equation}
It should also be noted that every time series sequence has components such as seasonality \& trend , these characteristics are also important when predicting future estimates, in this case the covariates $X_{t+\omega}^{(i)}~(\mathrm{here}~\omega = 1,...,\Omega~\mathrm{\&}~i = 1, ..., N)$ which provides additional info for the estimation framework in Equation~\ref{direct}. Finally the entire joint distribution of predicted estimates are provided in Equation.~\ref{direct-cov}.

\begin{equation}\label{direct-cov}
    P\left({\mathbf y}_{(t+1):(t+\Omega)}|{\mathbf y}_{1:t}\right) =
    \prod\limits_{\omega=1}^{\Omega} p({\mathbf y}_{t+\omega}|{\mathbf y}_{1:t}, X_{t+\omega}^{(i)}, i=1,...,N).
\end{equation}

Therefore the key challenge DeepTCN is able to answer is its ability to provide a suitable framework which includes past datapoints ${\mathbf y}_{1:t}$ and covariates $X_{t+\omega}^{(i)}$ to provide better estimates.

%%%%%%%%%%%%%%%%%%%%%%%%%%%%%%%%%%%%%%%%%%%%%%%%%%%%%%%%%%%%%%%%%%

%%%%%%%%%%%%%%%%%%%%%%%%%%%%%%%%%%%%%%%%%%%%%%%%%%%%%%%%%%%%%%%%%%%%

%%%%%%%%%%%%%%%%%%%%%%%%%%Deep AR%%%%%%%%%%%%%%%%%

\begin{figure}
\begin{center}
	\centerline{\includegraphics[width=\columnwidth]{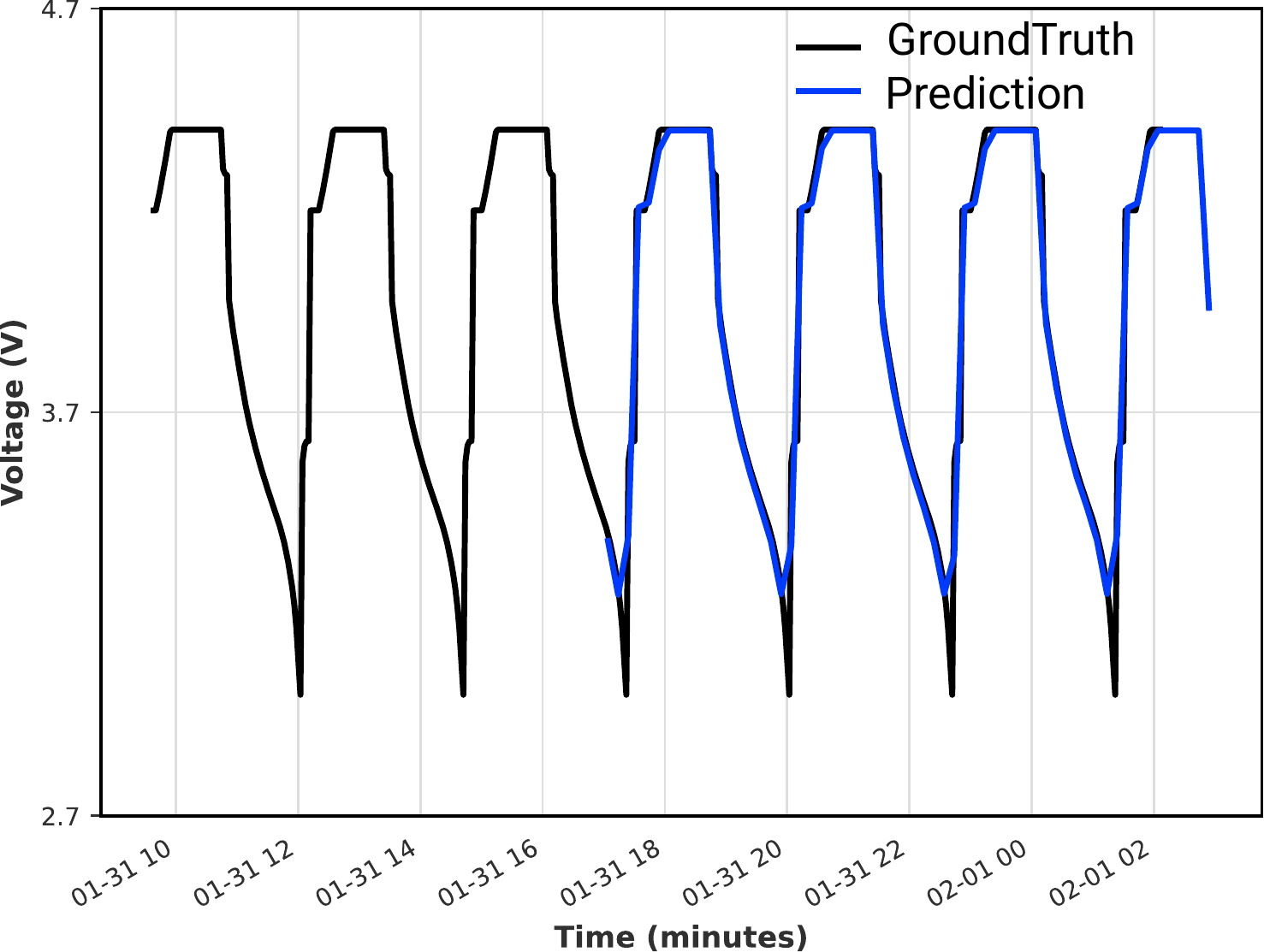}}
	\caption{DeepAR Voltage estimates with horizon 30}
	\label{fig:vol_deepAR}
\end{center}
\end{figure} 

\begin{figure}
\begin{center}
	\centerline{\includegraphics[width=\columnwidth]{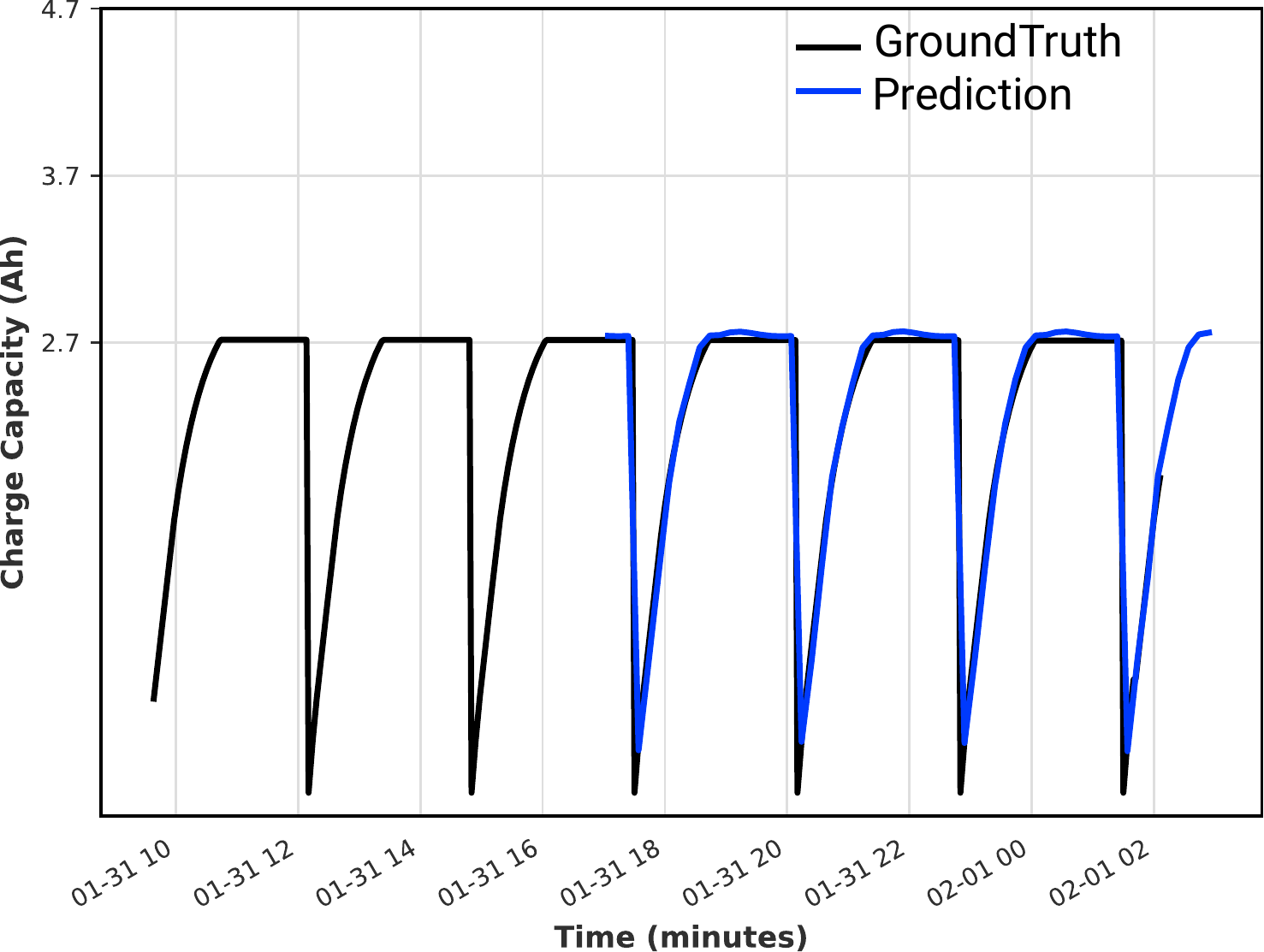}}
	\caption{DeepAR battery Capacity estimates with horizon 30}
	\label{fig:capacity_deepAR}
\end{center}
\end{figure} 

%%%%%%%%%%%%%%%%%%%%%%%%%%%%%%%%N-Beats%%%%%%%%%%%%%%%%%%

\begin{figure}
\begin{center}
	\centerline{\includegraphics[width=\columnwidth]{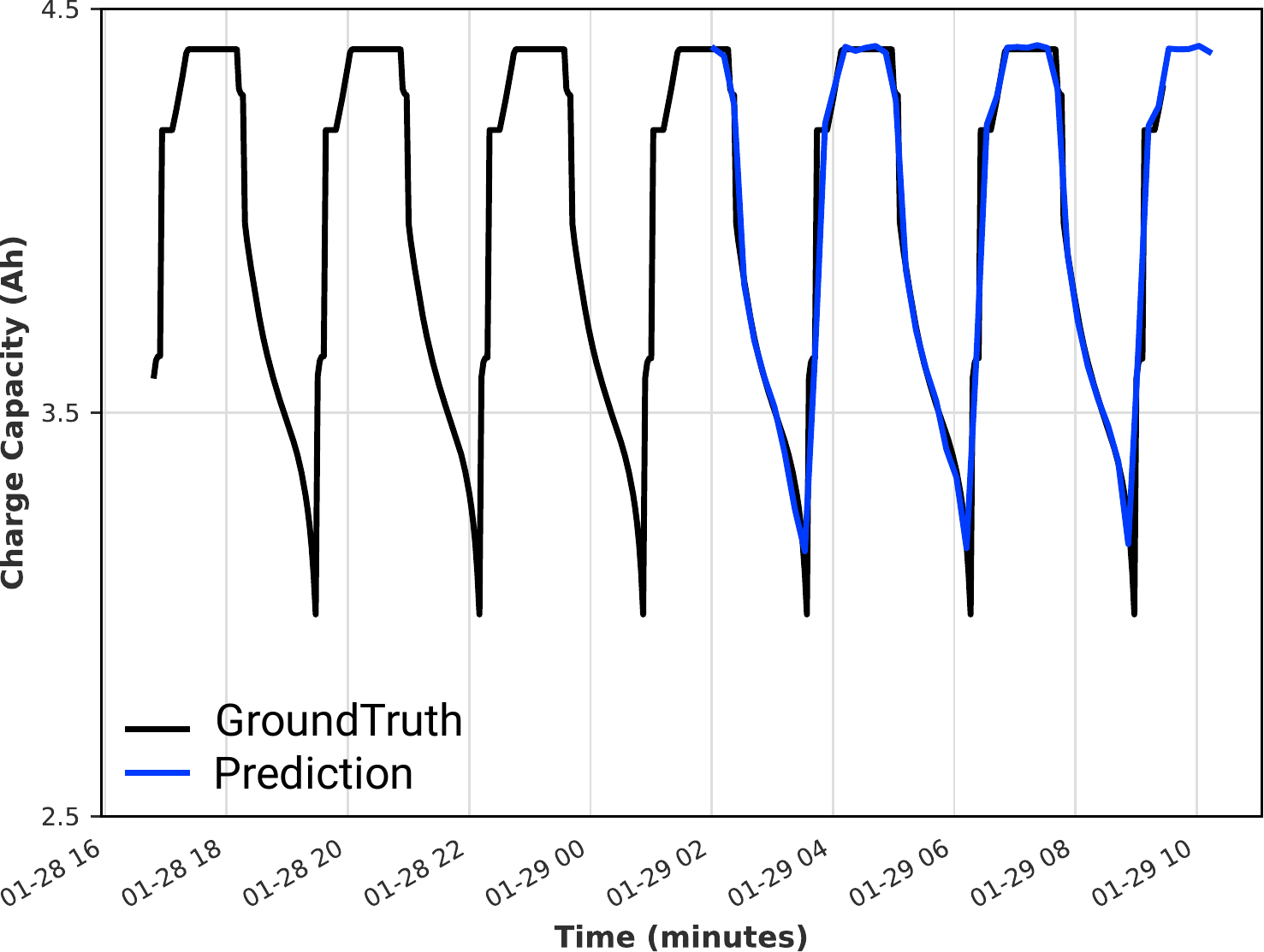}}
	\caption{N-BEATS Voltage estimates with horizon 30}
	\label{fig:vol_nbeats}
\end{center}
\end{figure} 

\begin{figure}
\begin{center}
	\centerline{\includegraphics[width=\columnwidth]{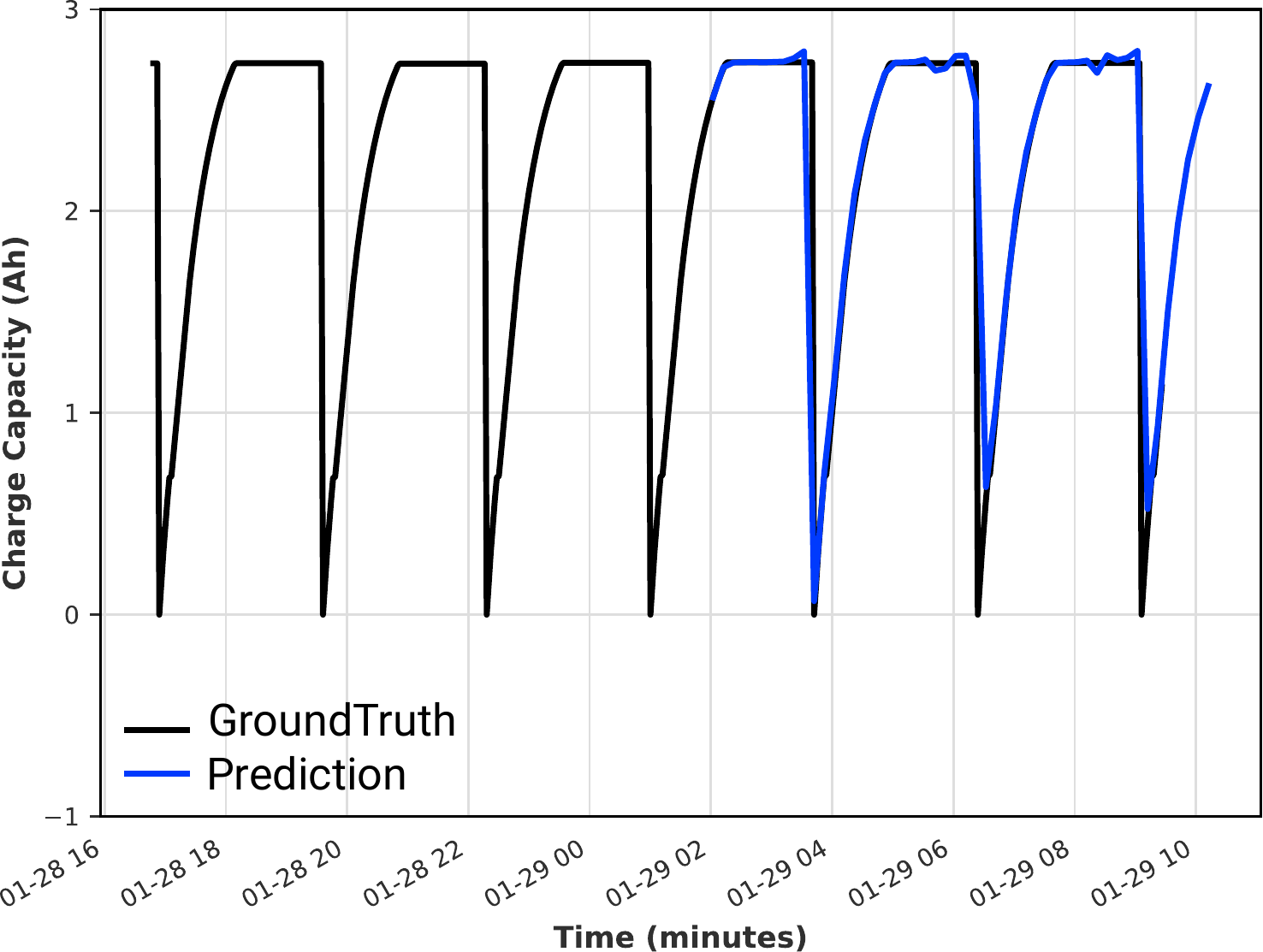}}
	\caption{N-BEATS battery Capacity estimates with horizon 30}
	\label{fig:capacity_nbeats}
\end{center}
\end{figure}

%%%%%%%%%%%%%%%%%%%%%%%%%%%%%%%%DeepTCN%%%%%%%%%%%%%%%%%%

\begin{figure}
\begin{center}
	\centerline{\includegraphics[width=\columnwidth]{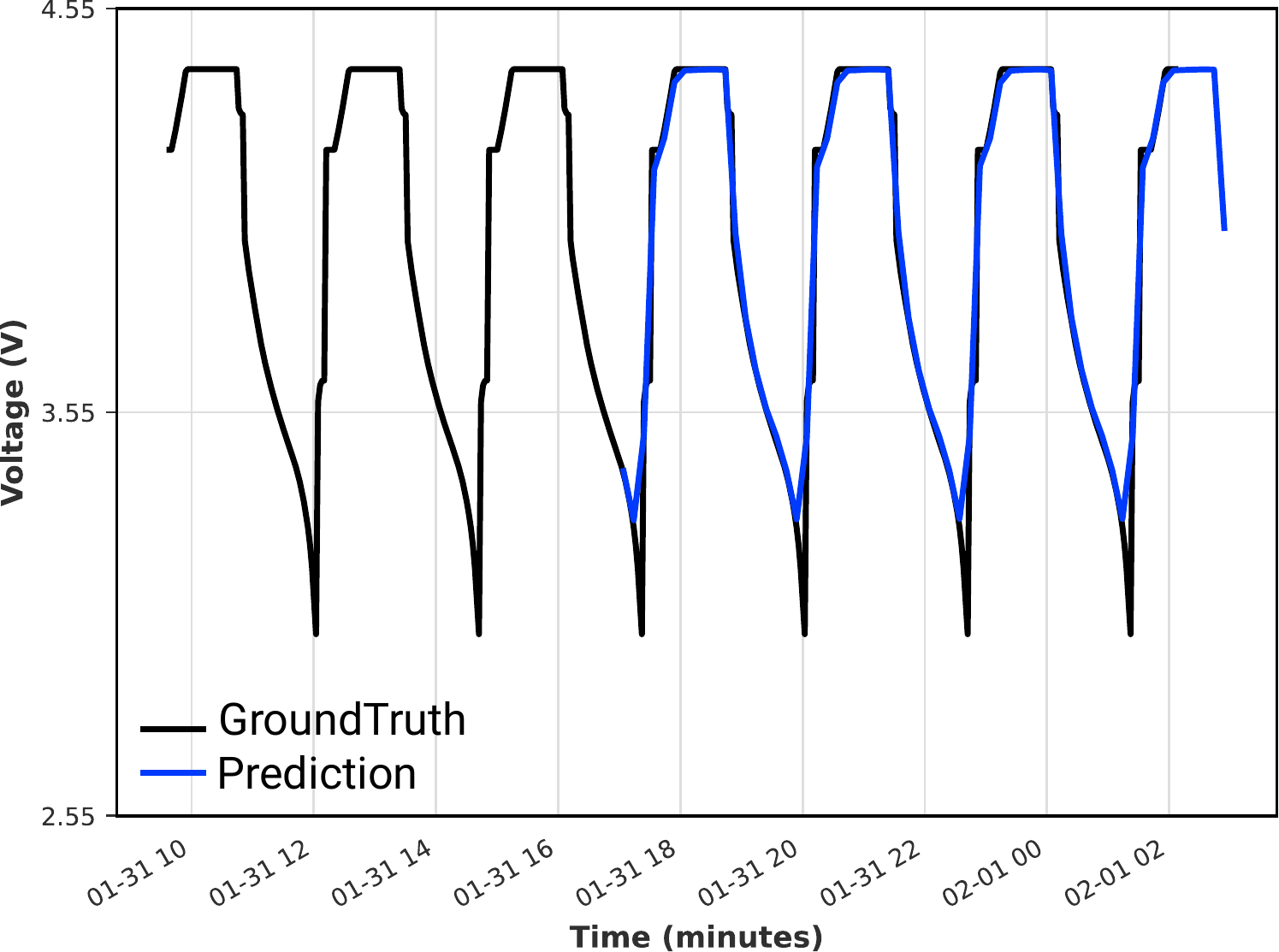}}
	\caption{DeepTCN Voltage estimates with horizon 30}
	\label{fig:vol_deeptcn}
\end{center}
\end{figure} 

\begin{figure}
\begin{center}
	\centerline{\includegraphics[width=\columnwidth]{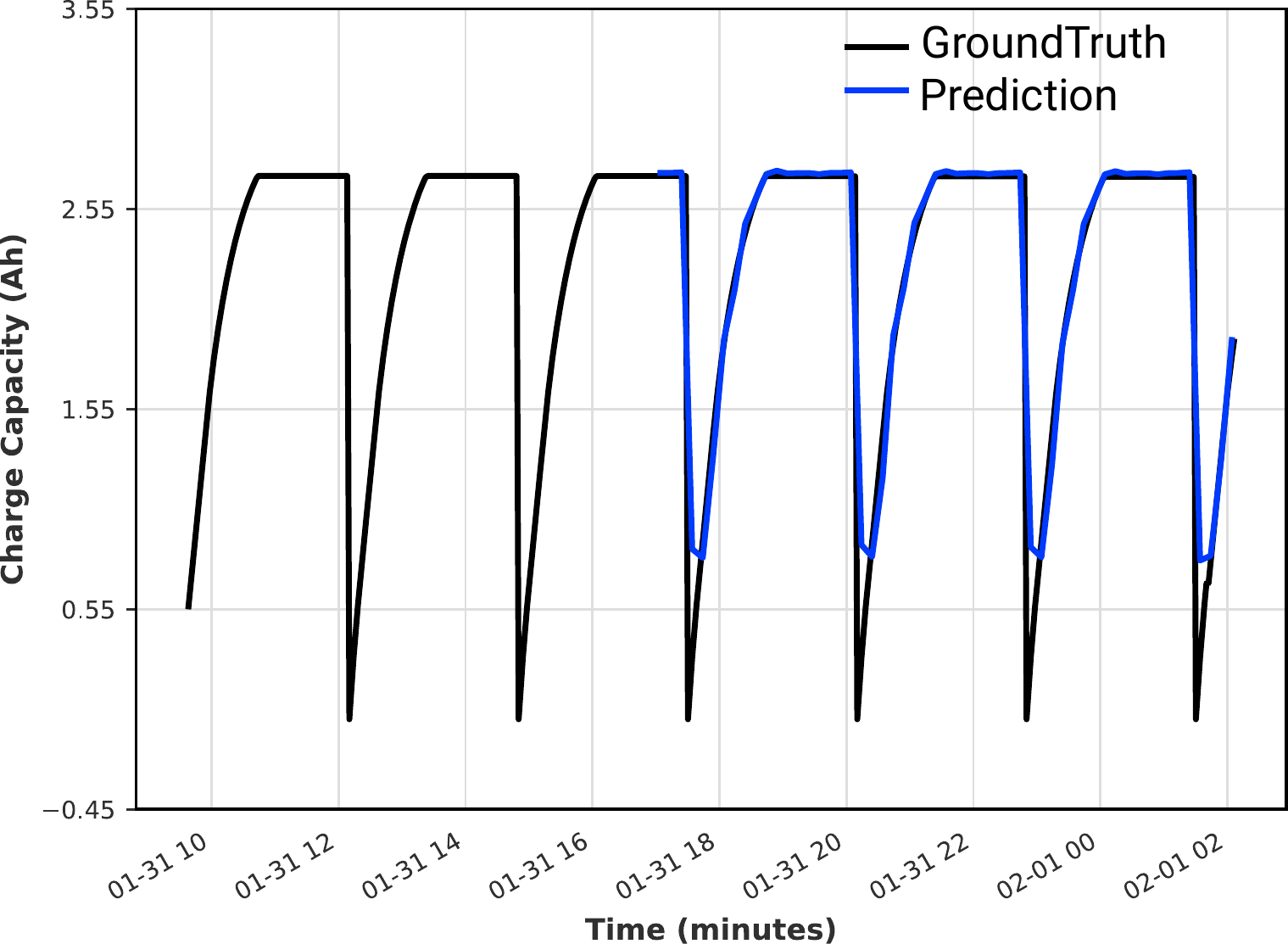}}
	\caption{DeepTCN battery Capacity estimates with horizon 30}
	\label{fig:capacity_deeptcn}
\end{center}
\end{figure} 

%%%%%%%%%%%%%%%%%%%%%%%%%%Error metrics%%%%%%%%%%%%%%%%%%
\begin{figure*}
\begin{center}
	\centerline{\includegraphics[scale=0.47]{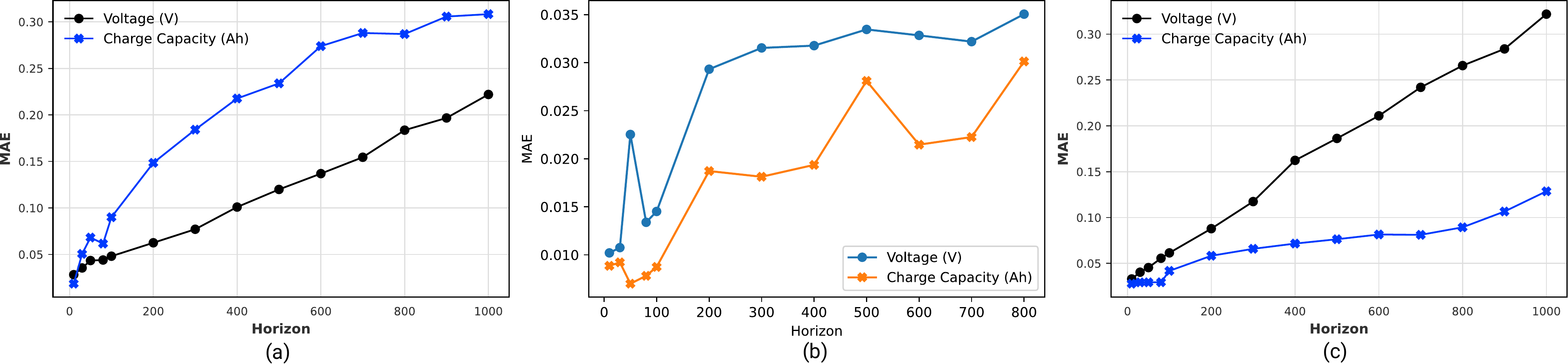}}
	\caption{Mean Absolute Error (MAE) metrics obtained for Voltage and battery capacity data \cite{calce} for various horizon values for DeepAR (a), N-BEATS (b) and DeepTCN (c)}
	\label{fig:dataset4_error}
\end{center}
\end{figure*} 

\begin{figure*}
\begin{center}
	\centerline{\includegraphics[scale=0.47]{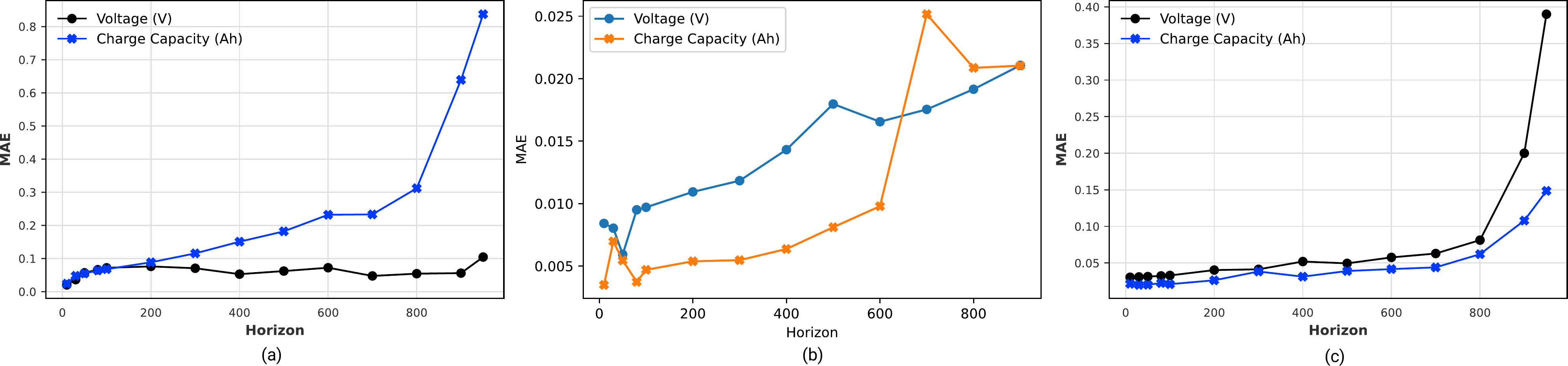}}
	\caption{Mean Absolute Error (MAE) metrics obtained for Voltage and battery capacity data \cite{pana} for various horizon values for DeepAR (a), N-BEATS (b) and DeepTCN (c)}
	\label{fig:mendeley}
\end{center}
\end{figure*}

\section{Results and Discussion}

All the deep learning models described in the paper were used to produce synthetic data, we employed two publicly available datasets to evaluate all the estimation techniques described here. The implementation details can be found here: \url{https://github.com/vageeshmaiya/Deep-Learning-based-Battery-Synthetic-Data}. After the preprocessing step of data normalization, we focussed on battery voltage and battery capacity from a $LiCoO_2$ battery \cite{calce} and a 18650 Panasonic battery \cite{pana}. The initial estimates of both voltage and capacity for DeeAR model can be found in Fig.~\ref{fig:vol_deepAR} and Fig.~\ref{fig:capacity_deepAR}. Subsequently we employed similar dataset to evaluate the N-BEATS and DeepTCN models as described in Fig.~\ref{fig:vol_nbeats}, Fig.~\ref{fig:capacity_nbeats}, Fig.~\ref{fig:vol_deeptcn} and Fig.~\ref{fig:capacity_deeptcn} respectively. We have used Mean Absolute Error (MAE) metrics described in Equation.~\ref{eq:mae} to observe the behaviour of these deep learning models over a range of horizon values as shown in Fig.~\ref{fig:dataset4_error} and Fig.~\ref{fig:mendeley}.

\begin{equation}\label{eq:mae}
\textrm{MAE}=\frac{1}{n}\sum_{i=1}^{n}|\mb{y}-\hat{\mb{y}}|    
\end{equation}

In Equation.~\ref{eq:mae}, $\mb{y}$ represents ground truth values \& the estimated values are shown by $\hat{\mb{y}}$. We can observe that DeepTCN provides the best performance across both datasets and for both battery voltage and battery capacity estimates for a varied horizon length. In general, battery voltage estimate error metrics increase overtime due to a lower voltage resolution of 0.1 V across different battery datasets. From this exhaustive analysis, we can consider deploying DeepTCN to provide high-fidelity synthetic battery datasets during sparse dataset scenarios.

\section{Conclusion}
Lack of abundant labelled battery dataset is a cause of concern when researchers are developing novel battery capacity and health estimation algorithms backed by deep learning. Present open access datasets do not contain the necessary diversity (various charge/discharge profiles) to develop generalized estimation algorithms. The duration and the hardware cost involved in collecting battery measurements are also some of the major reasons for data scarcity. This paper compares and evaluates three state-of-the-art deep learning approaches to generate high quality battery synthetic datasets. All approaches were evaluated on two publicly available datasets and it was found that using Temporal Convolutional Neural Network provided the best performance for such data augmentation purposes.

%\clearpage 
%\newpage

\end{document}